\documentclass{article}
\usepackage{spconf,amsmath,graphicx,hyperref}
\usepackage{amsmath,amssymb,graphicx,booktabs}
\usepackage{adjustbox}
\usepackage{multirow}
\usepackage{booktabs, array}
\usepackage{caption}
\usepackage[table]{xcolor}
\definecolor{first}{RGB}{191, 225, 201} 
\definecolor{second}{RGB}{227, 237, 185} 
\definecolor{third}{RGB}{254, 250, 194} 

\title{Geodesic Prototype Matching via Diffusion Maps for Interpretable Fine-Grained Recognition}
\name{Junhao Jia$^{1,*}$, Yunyou Liu$^{1}$, Yifei Sun$^{2}$, Huangwei Chen$^{2}$, Feiwei Qin$^{1,*}$\thanks{This work was supported in part by National Undergraduate Training Program for Innovation and Entrepreneurship (No.202510336076), Zhejiang Students‘ Technology and Innovation Program (No.GK250701201018 and No.GK250701201041), Fundamental Research Funds for the Provincial Universities of Zhejiang (No. GK259909299001-006), Anhui Provincial Joint Construction Key Laboratory of Intelligent Education Equipment and Technology (No. IEET202401), and the ‘Pioneer’ and ‘Leading Goose’ R\&D Program of Zhejiang (No.2025C04001).}, Changmiao Wang$^{3}$, Yong Peng$^{1}$}

\address{$^{1}$ Hangzhou Dianzi University, Hangzhou, China \\
         $^{2}$ Zhejiang University, Hangzhou, China \\
         $^{3}$ Shenzhen Research Institute of Big Data, Shenzhen, China \\
         $^{*}$ Corresponding author: 23080631@hdu.edu.cn, qinfeiwei@hdu.edu.cn
}

\begin{document}
%
\maketitle
\begin{abstract}
Nonlinear manifolds are pervasive in deep visual features, where Euclidean distances can misrepresent true similarity. This mismatch is particularly detrimental to prototype-based interpretable fine-grained recognition, where even subtle semantic distinctions are crucial. To mitigate this issue, this work presents a novel paradigm for prototype-based recognition by grounding similarity in the intrinsic geometry of deep features. Concretely, we distill the latent manifold structure of each class into a diffusion space and, critically, devise a differentiable Nystr\"{o}m interpolation to make this geometry accessible to both unseen samples and learnable prototypes. To maintain efficiency, we employ compact per-class landmark sets with periodic updates. This strategy keeps the embedding synchronized with the evolving backbone, enabling fast inference at scale. Comprehensive experiments on two benchmark datasets demonstrate that our GeoProto yields prototypes focusing on semantically corresponding parts, significantly outperforming Euclidean prototype networks.
\end{abstract}
\section{Introduction}
\label{sec:intro}
\begin{figure}[t]
\centering
\includegraphics[width=0.9\columnwidth]{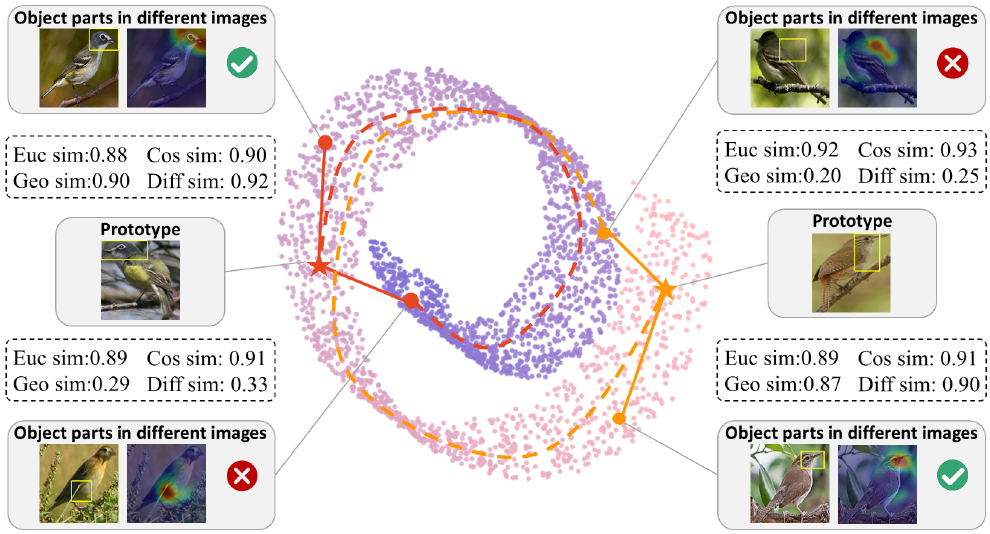}
\caption{Diffusion (geodesic) similarity respects the class-manifold, avoiding Euclidean “shortcuts” and yielding semantically consistent prototype–part matches.}
\label{fig1}
\end{figure}
Prototype learning~\cite{chen2019looks} offers an intrinsically interpretable paradigm for image recognition, in which a model makes predictions by learning prototypical patches and aggregating similarity scores to produce class logits, yielding case-based explanations grounded in visual evidence. However, most existing methods~\cite{wang2023learning} evaluate similarity using Euclidean distance in the feature space, which implicitly assumes that the space is globally flat~\cite{hoffmann2021looks}. In practice, images of the same class can lie on a high-dimensional manifold where the straight-line distance often overestimates the dissimilarity across images~\cite{tenenbaum2000global}. As depicted in Fig.~\ref{fig1}, Euclidean similarity collapses manifold geometry into straight lines, creating shortcut neighbors and diverting queries to inferior prototypes. The core issue is that Euclidean similarity overlooks the underlying manifold structure of the feature space~\cite{coifman2006diffusion, jia2025brain, sun2025wdt}.

\begin{figure*}[t]
\centering
\includegraphics[width=0.9\textwidth]{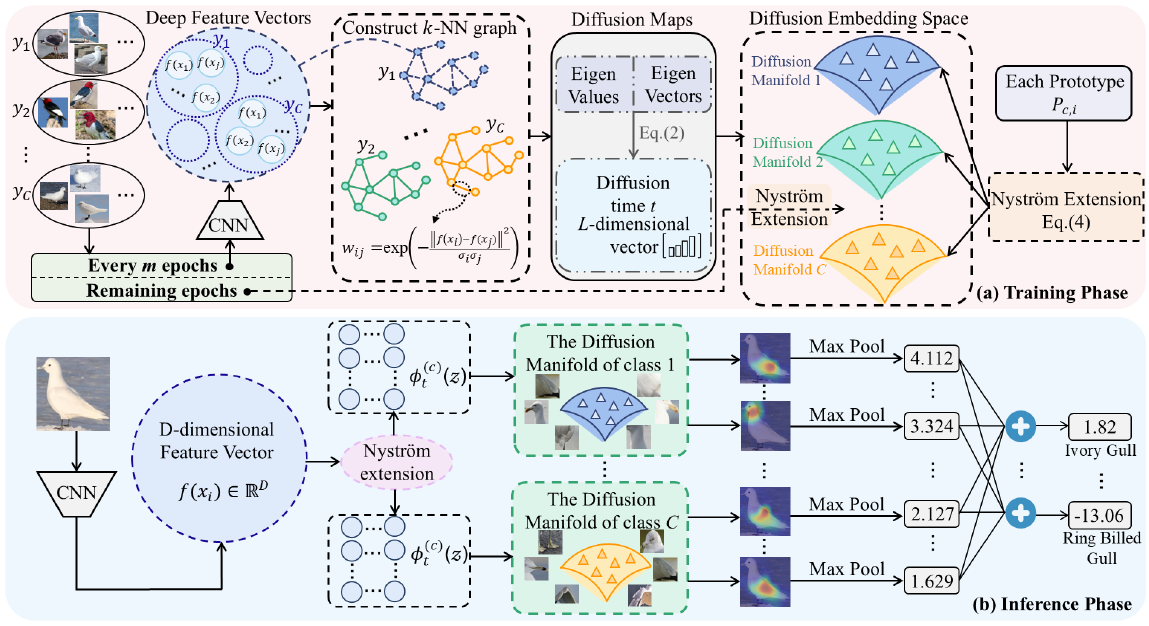}
\caption{The overview of our proposed GeoProto framework.
(a) Training: build class-wise diffusion (geodesic) manifolds from CNN features and embed prototypes via Nyström.
(b) Inference: map a query into each class manifold via Nyström, compute geodesic similarity to prototypes, then max-pool and aggregate to produce the class score and prototype–part explanations.}
\label{fig2}
\end{figure*}

In response, we introduce GeoProto, a novel geodesic prototype matching framework. It aligns queries to prototypes along geodesic paths, thereby uncovering the underlying structure across samples. To the best of our knowledge, this is the first work to systematically revisit and replace the distance paradigm in prototype-based interpretability methods. From this perspective, our primary contributions are:

(1) We identify that Euclidean similarity misaligns with class manifolds and recast prototype reasoning with a geodesic metric, providing a manifold-aware notion of similarity.

(2) We propose an end-to-end differentiable framework that learns prototypes and matches them on the manifold via diffusion distance with Nystr\"{o}m Extension, thus producing faithful case-based explanations.

(3) Comprehensive experiments on two benchmark datasets demonstrate that GeoProto outperforms Euclidean-based prototype networks in both accuracy and interpretability.

\section{Method}
\label{sec:Method}

As illustrated in Fig.~\ref{fig2}, we propose the GeoProto framework, which replaces Euclidean similarity with diffusion-based geodesic similarity for prototype matching, enabling manifold-aware prototypical learning and inference.

\subsection{Class-Wise Graph Construction with Local Scaling}
Let $\{(x_i, y_i)\}_{i=1}^{N}$ be the training set of images with class labels $y_i \in \{1,\dots,C\}$. We first extract deep feature vectors $f(x_i) \in \mathbb{R}^D$ from a CNN backbone (e.g., a ResNet or VGG). For each class $c$, we construct an affinity graph $G_c$ over that class's training samples to model the local manifold structure. Each node represents a feature $f(x_i)$ with $y_i = c$. We connect each sample to its $k$ nearest neighbors ($k$-NN) within the same class in Euclidean feature space to form the graph edges. 

We assign edge weights using a Gaussian kernel with local scaling, for an edge between node $i$ and $j$, the weight is: 
\begin{equation}
    w_{ij} = \exp\!\Big(-\frac{\|f(x_i) - f(x_j)\|^2}{\sigma_i\, \sigma_j}\Big),
\end{equation}
where $\sigma_i$ is an adaptive bandwidth for node $i$, set to the Euclidean distance from $f(x_i)$ to its $k$-th nearest neighbor in class $c$ (similarly for $\sigma_j$). This local scaling ensures the affinity is normalized by local density, making the graph similarity more robust across dense and sparse regions. We symmetrize $w_{ij}=w_{ji}$ and define the degree $d_i = \sum_j w_{ij}$. The class-$c$ affinity matrix is $\mathbf{W}^{(c)} \in \mathbb{R}^{n_c \times n_c}$ (where $n_c$ is the number of training samples of class $c$), and the corresponding row-normalized transition matrix is $\mathbf{P}^{(c)} = (\mathbf{D}^{(c)})^{-1} \mathbf{W}^{(c)}$, with $\mathbf{D}^{(c)} = \mathrm{diag}(d_1,\dots,d_{n_c})$. $\mathbf{P}^{(c)}$ defines a Markov random walk on the feature graph of class $c$.

\begin{table*}[t]
\centering
\caption{Classification accuracy (\%) on two benchmark datasets. Best results are highlighted as \colorbox{first}{first}, \colorbox{second}{second} and \colorbox{third}{third}.}
\setlength{\tabcolsep}{0pt}
\fontsize{9.5pt}{10pt}\selectfont
\newlength{\MethodColW}\setlength{\MethodColW}{2.5cm}
\newlength{\NumColW}\setlength{\NumColW}{\dimexpr(\textwidth-\MethodColW)/12\relax}

\begin{tabular}{
  >{\centering\arraybackslash}p{\MethodColW}%
  *{12}{>{\centering\arraybackslash}p{\NumColW}}%
}
\toprule
\multirow{2.4}{*}{Method}
& \multicolumn{6}{c}{\textbf{CUB-200-2011~\cite{wah2011caltech}}}
& \multicolumn{6}{c}{\textbf{Stanford Cars~\cite{krause20133d}}} \\
\cmidrule(lr){2-7} \cmidrule(lr){8-13}
& V16 & V19 & R34 & R50 & D121 & D161
& V16 & V19 & R34 & R50 & D121 & D161 \\
\midrule
Baseline & 70.1 & 70.8 & 75.4 & 78.1 & 77.3 & 79.2
         & 80.5 & 81.2 & 83.1 & 85.0 & \cellcolor{third}{84.4} & 85.3 \\
\midrule
ProtoPNet~\cite{chen2019looks}      & 69.4 & 71.8 & 71.5 & 80.3 & 73.1 & 74.9
                                & 79.6 & 80.4 & 80.7 & 84.5 & 81.2 & 82.1 \\
TesNet~\cite{wang2021interpretable} & 74.8 & 76.7 & 75.4 & 85.4 & 78.3 & 78.7
                                & 81.8 & 82.5 & 81.8 & \cellcolor{third}{87.1} & 83.4 & 83.1 \\                       
ProtoPool~\cite{rymarczyk2022interpretable} & 74.3 & 74.7 & 75.4 & 82.6 & 77.5 & 79.7
                           & 80.3 & 81.1 & 81.1 & 84.6 & 82.4 & 83.2 \\
ProtoKNN~\cite{ukai2023this}  & 76.2 & 76.9 & 77.1 & 80.1 & 78.9 & 80.6
                           & \cellcolor{third}{83.5} & \cellcolor{second}{84.1} & 83.9 & 85.3 & \cellcolor{second}{85.3} & \cellcolor{third}{85.7} \\

ProtoConcepts~\cite{ma2023looks} & 69.8 & 71.5 & 72.9 & 76.8 & 73.7 & 74.4
                                   & 80.6 & 81.5 & 82.1 & 84.1 & 82.4 & 83.0 \\

ST-ProtoPNet~\cite{wang2023learning} & 75.3 & 76.9 & 76.8 & 83.9 & 77.6 & 79.3
                                & 81.7 & 82.8 & 82.5 & 85.8 & 82.9 & 83.7 \\
SDFA-SA~\cite{huang2023evaluation} & 75.7 & 77.1 & 76.8 & \cellcolor{third}{85.6} & \cellcolor{second}{80.3} & 79.9
                                & 81.6 & 82.3 & 81.8 & 86.6 & 84.2 & 83.4 \\
MGProto~\cite{wang2025mixture} & \cellcolor{second}{78.8} & \cellcolor{second}{79.0} & \cellcolor{third}{80.1} & \cellcolor{second}{86.2} & \cellcolor{second}{80.3} & \cellcolor{third}{82.1}
                                & 83.4 & 83.4 & \cellcolor{third}{84.4} & \cellcolor{second}{87.2} & 84.2 & \cellcolor{third}{85.7} \\
CBC~\cite{saralajew2025robust}
& \cellcolor{third}{78.3} & 78.6 & \cellcolor{second}{80.3} & 85.5 & \cellcolor{third}{79.6} & \cellcolor{second}{82.3}
& \cellcolor{second}{83.6} & \cellcolor{third}{83.7} & \cellcolor{second}{84.9} & 86.9 & 83.9 & \cellcolor{second}{86.0} \\
ProtoArgNet~\cite{ayoobi2025protoargnet}
& 77.9 & \cellcolor{third}{78.7} & 79.2 & 85.1 & 79.1 & 81.5
& 83.0 & 82.9 & 84.2 & 85.0 & 84.0 & 84.3 \\
\midrule
\textbf{GeoProto (Ours)}
& \cellcolor{first}{\textbf{80.5}} & \cellcolor{first}{\textbf{80.1}} & \cellcolor{first}{\textbf{82.1}} & \cellcolor{first}{\textbf{87.8}} & \cellcolor{first}{\textbf{81.2}} & \cellcolor{first}{\textbf{84.3}}
& \cellcolor{first}{\textbf{85.1}} & \cellcolor{first}{\textbf{85.1}} & \cellcolor{first}{\textbf{86.4}} & \cellcolor{first}{\textbf{88.9}} & \cellcolor{first}{\textbf{85.7}} & \cellcolor{first}{\textbf{87.4}} \\
\bottomrule
\label{Table1}
\end{tabular}
\end{table*}

\subsection{Diffusion Maps Embedding and Nystr\"{o}m Extension}
We apply the Diffusion Maps~\cite{coifman2006diffusion} on each class graph $G_c$ to obtain an embedding that encapsulates the manifold geometry. Specifically, we perform an eigendecomposition of $\mathbf{P}^{(c)}$. Let $\{\lambda_{\ell}^{(c)}, \psi_{\ell}^{(c)}(\cdot)\}_{\ell=0}^{L}$ be the eigenvalues and eigenvectors (eigenfunctions) of $\mathbf{P}^{(c)}$, indexed in decreasing order $1=\lambda_0 > \lambda_1 \ge \lambda_2 \dots \ge \lambda_L$. We ignore the trivial eigenvector $\psi_0$. The diffusion map of a training sample $x_i$ in class $c$ at diffusion time $t$ is given by the $L$-dimensional vector:
\begin{equation}\label{eq:diffmap}
    \Phi_t^{(c)}(x_i) = \big(\lambda_1^t\, \psi_1^{(c)}(i),\; \lambda_2^t\, \psi_2^{(c)}(i),\; \dots,\; \lambda_L^t\, \psi_L^{(c)}(i)\big).
\end{equation}
In the diffusion embedding space, the Euclidean distance between two points corresponds exactly to their diffusion distance, which aggregates $t$-step connectivity on the graph. In particular, the squared diffusion distance can be expressed as: 

\begin{equation}\label{eq:diffdist}
d_{\mathit{diff},t}^{(c)}(x_i, x_j)^2 \;=\; \sum_{\ell=1}^{L} (\lambda_{\ell}^{(c)})^{2t}\,\big(\psi_{\ell}^{(c)}(i) - \psi_{\ell}^{(c)}(j)\big)^2,
\end{equation}
which converges to the geodesic distance on the underlying manifold as the sample density increases. Thus, diffusion distance provides a provably better approximation to true intrinsic distances than raw Euclidean distance on high-dimensional data.

To compute diffusion-based distances for a new test image or an arbitrary feature vector $z \in \mathbb{R}^D$ belonging to class $c$, we utilize the Nystr\"{o}m extension for out-of-sample embedding. Given the precomputed eigenpairs $\{ \lambda_{\ell}^{(c)}, \psi_{\ell}^{(c)}\}$ from training data of class $c$, we compute the affinity between $z$ and each training sample $x_j$ of class $c$: $k(z, x_j) = \exp(-\frac{\|z - f(x_j)\|^2}{\sigma(z)\, \sigma_j})$, where we define $\sigma(z)$ by the distance from $z$ to its $k$-NN among class-$c$ training features. Then, the Nystr\"{o}m-extended diffusion coordinates of $z$ are obtained by:
\begin{equation}\label{nystrom}
    \Phi_t^{(c)}(z) \;=\; \frac{1}{\mathbf{\lambda}^{(c)}} \sum_{j=1}^{n_c} \frac{k(z,x_j)}{d_j}\, \big(\lambda_1^t \psi_1^{(c)}(j), \dots, \lambda_L^t \psi_L^{(c)}(j)\big),
\end{equation}
where the division by $\mathbf{\lambda}^{(c)}$ denotes elementwise division by the vector of eigenvalues $[\lambda_1^{(c)},\dots,\lambda_L^{(c)}]$. This effectively interpolates $z$ into the diffusion space spanned by the training data. The mapping $z \mapsto \Phi_t^{(c)}(z)$ is smooth since it is composed of Gaussian kernel evaluations and linear combinations; thus, gradients can flow through the Nystr\"{o}m embedding step during training.

\subsection{Prototype Matching}

In prototype-based classification~\cite{chen2019looks,wang2021interpretable}, each class is endowed with $m$ learnable prototype vectors $p_{c,i}\in\mathbb{R}^D$ ($i=1,\ldots,m$). To keep them interpretable and on-manifold, we project prototypes during training and at inference. Specifically, we first map $p_{c,i}$ into the class-$c$ diffusion space via the Nystr\"om extension in Eq.~\ref{nystrom}, obtaining $\Phi_t^{(c)}(p_{c,i})$. We then anchor the prototype to the nearest training patch from class $c$ in diffusion coordinates, searching over the candidate set $\mathcal{S}_c$ extracted from convolutional feature maps:
\begin{equation}
\tilde p_{c,i}
\;=\;
\arg\min_{u\in \mathcal{S}_c}
\big\|
\Phi_t^{(c)}(p_{c,i})-\Phi_t^{(c)}(u)
\big\|_2 .
\end{equation}

\textbf{Inference.} For each class $c$, we apply the Nystr\"om extension associated with $G_c$ to the input feature $z=f(x)$ to obtain $\Phi_t^{(c)}(z)$. We then compare, within the same class-conditional diffusion space, $\Phi_t^{(c)}(z)$ against the projected prototypes of class $c$ to compute intra-class distances:
\begin{equation}
d_{c,i}(x)\;=\;\big\|\Phi_t^{(c)}(z)-\Phi_t^{(c)}(\tilde p_{c,i})\big\|_2 .
\end{equation}

Given intra-class diffusion distances $\{d_{c,i}(x)\}$, we follow prior prototype-aggregation schemes~\cite{chen2019looks}: distances are monotonically transformed into similarities and aggregated by a class-restricted nonnegative linear head to obtain logits.

\section{Experiments}
\label{sec:exp}
\subsection{Experimental Setup}
We evaluate GeoProto on CUB-200-2011~\cite{wah2011caltech} and Stanford Cars~\cite{krause20133d} with VGG-16/19, ResNet-34/50, and DenseNet-121/161 backbones. All models are pre-trained on ImageNet~\cite{deng2009imagenet}, except that ResNet-50 on CUB-200-2011 is pre-trained on iNaturalist~\cite{van2018inaturalist}. For a fair comparison, we fix all other settings and replace Euclidean matching with a class-conditional diffusion distance computed by a differentiable Nystr\"om extension. We follow the same evaluation protocol~\cite{wang2023learning,huang2023evaluation,wang2025mixture}, reporting Accuracy (ACC), Consistency (Cons.), Stability (Stabil.), OIRR, DAUC, and ECE.


\subsection{Comparsion with SOTA Methods}
As presented in Table~\ref{Table1}, GeoProto consistently surpasses all prototype-based interpretable models. On CUB-200-2011 with ResNet-50, GeoProto attains 87.8\% accuracy, a 1.6\% gain over MGProto 86.2\%; on Stanford Cars with ResNet-50, GeoProto reaches 88.9\%, again leading all baselines. As evident, GeoProto delivers the best accuracy across all entries in Table~\ref{Table1}, spanning both datasets and all backbones.
\begin{figure}[!t]
  \centering
  \includegraphics[width=\columnwidth,keepaspectratio]{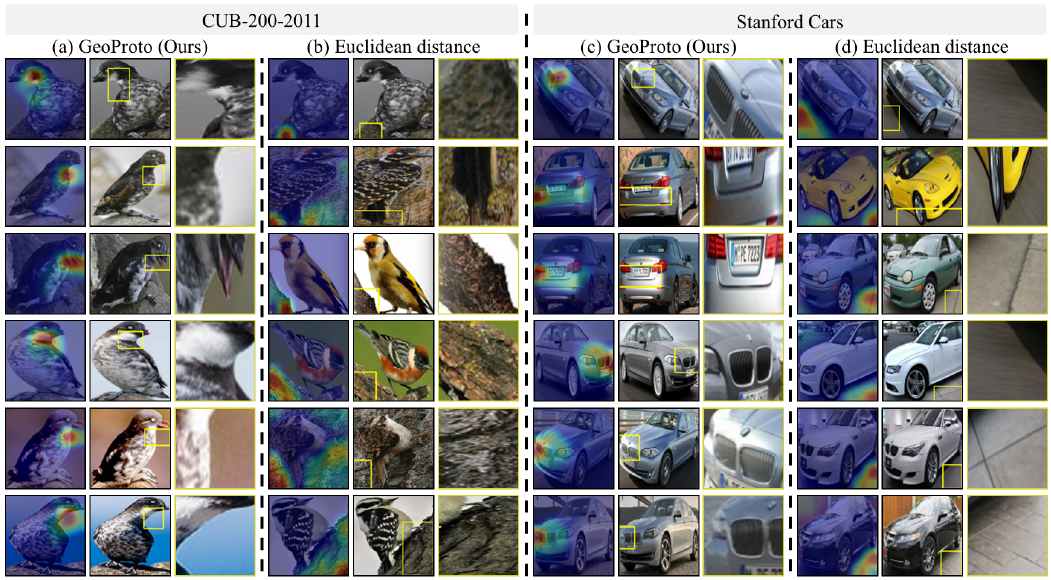}
  \caption{Each panel shows a prototype and its five nearest patches. GeoProto yields more class-consistent parts, while Euclidean tends to include off-manifold textures.}
  \label{fig3}
\end{figure}

\begin{table}[t]
\caption{Ablation of distance metric and normalization on CUB-200-2011 with ResNet-50.}
\centering
\fontsize{9pt}{10pt}\selectfont
\setlength{\tabcolsep}{1.5pt}
\begin{tabular}{ccccccccccc}
\toprule
Metric & Norm & $t$ & $L$ & ACC & OIRR & DAUC & Cons. & Stabil. \\
\midrule
Euclidean & -- & -- & -- & 80.3 & 23.92 & 3.95 & 92.93 & 45.63 \\
Cosine    & -- & -- & -- & 81.7 & 23.65 & 3.62 & 93.03 & 46.12 \\
Diag-Mahalanobis & -- & -- & -- & 82.3 & 23.33 & 3.35 & 93.23 & 46.43 \\
\midrule
Diffusion Maps & None   & 4 & 32 & \cellcolor{third}{86.9} & \cellcolor{third}{22.83} & \cellcolor{third}{3.13} & \cellcolor{third}{93.34} & \cellcolor{third}{46.84} \\
Diffusion Maps & Energy & 4 & 32 & \cellcolor{second}{87.6} & \cellcolor{second}{22.14} & \cellcolor{second}{3.04} & \cellcolor{second}{93.54} & \cellcolor{second}{47.13} \\
\textbf{Diffusion Maps} & \textbf{ZCA} & \textbf{4} & \textbf{32} & \cellcolor{first}{\textbf{87.8}} & \cellcolor{first}{\textbf{21.84}} & \cellcolor{first}{\textbf{2.92}} & \cellcolor{first}{\textbf{93.74}} & \cellcolor{first}{\textbf{47.44}} \\
\bottomrule
\end{tabular}
\label{tab:ablation-metric}
\end{table}

\begin{table}[t]
\caption{Ablation of graph construction and diffusion parameters on CUB-200-2011 with ResNet-50.}
\centering
\fontsize{9pt}{10pt}\selectfont
\setlength{\tabcolsep}{1.4pt}
\begin{tabular}{cccccccccccc}
\toprule
$k$ & Local & $t$ & $L$ & Comp. & A.P. & ACC & ECE & OIRR & DAUC & Cons. & Stabil. \\
\midrule
10 & off & 2 & 16 & 2.8 & 7.4 & 85.4 & 3.27 & 23.24 & 3.31 & 93.12 & 46.37 \\
10 & on  & 2 & 16 & 1.6 & 6.3 & \cellcolor{third}{86.1} & 3.03 & 23.02 & 3.24 & \cellcolor{third}{93.26} & \cellcolor{third}{46.58} \\
20 & off & 4 & 32 & \cellcolor{third}{1.4} & \cellcolor{third}{5.9} & \cellcolor{second}{86.9} & \cellcolor{third}{2.91} & \cellcolor{third}{22.83} & \cellcolor{second}{3.13} & \cellcolor{second}{93.34} & \cellcolor{second}{46.84} \\
\textbf{20} & \textbf{on} & \textbf{4} & \textbf{32} & \cellcolor{second}{\textbf{1.1}} & \cellcolor{second}{\textbf{5.2}} & \cellcolor{first}{\textbf{87.8}} & \cellcolor{first}{\textbf{2.59}} & \cellcolor{first}{\textbf{21.84}} & \cellcolor{first}{\textbf{2.92}} & \cellcolor{first}{\textbf{93.74}} & \cellcolor{first}{\textbf{47.44}} \\
30 & on  & 8 & 64 & \cellcolor{first}{1.0} & \cellcolor{first}{4.8} & \cellcolor{third}{86.1} & \cellcolor{second}{2.73} & \cellcolor{second}{22.61} & \cellcolor{third}{3.17} & 93.18 & 46.52 \\
\bottomrule
\end{tabular}
\label{Table3}
\end{table}

\begin{table}[t]
\centering
\caption{Ablation of Nyström landmarks selection and update frequency on Stanford Cars with ResNet-50.}
\label{Table4}
\fontsize{9pt}{10pt}\selectfont
\setlength{\tabcolsep}{1.0pt}

\begin{tabular}{ccccccc}
\toprule
Selection & Pool & Landmarks & Update Freq. & ACC & ECE & Lat.(ms) \\
\midrule
Random    & Per-Class & 256  & Fixed         & 87.7 & 2.9 & \cellcolor{first}{4.9} \\
Random    & Per-Class & 512  & Every 20 Ep.  & \cellcolor{third}{87.9} & \cellcolor{third}{2.8} & \cellcolor{third}{5.4} \\
K-Means   & Per-Class & 512  & Every 20 Ep.  & \cellcolor{second}{88.4} & \cellcolor{second}{2.7} & \cellcolor{second}{5.2} \\
K-Means   & Per-Class & 1024 & Fixed         & \cellcolor{second}{88.4} & \cellcolor{second}{2.7} & 6.0 \\
K-Means   & Global    & 8000 & Fixed         & \cellcolor{second}{88.4} & \cellcolor{second}{2.7} & 6.8 \\
\textbf{K-Means} & \textbf{Per-Class} & \textbf{768} & \textbf{Every 20 Ep.} & \cellcolor{first}{\textbf{88.9}} & \cellcolor{first}{\textbf{2.6}} & \textbf{5.6} \\
\bottomrule
\end{tabular}

\end{table}

\subsection{Visualization Protocol}

For each prototype, GeoProto ranks same-class patches using diffusion distance computed via Nyström extension on class landmarks, while in the original feature space we rank by Euclidean distance. As shown in Fig.~\ref{fig3}, GeoProto preferentially localizes semantically coherent parts, whereas Euclidean distance systematically accentuates background or edge textures, in line with the observed quantitative improvements. 

\subsection{Ablation Study}
\textbf{Effect of Distance Metric and Normalization.} On CUB-200-2011 with ResNet-50, replacing Euclidean with cosine or diagonal Mahalanobis improves accuracy from 80.3\% to 81.7\% and 82.3\%. Diffusion Maps with $t=4$ and $L=32$ yields 86.9\% without normalization, 87.6\% with energy normalization and 87.8\% with ZCA, while OIRR and DAUC drop to 21.84 and 2.92, as shown in Table~\ref{tab:ablation-metric}. These results indicate that diffusion distance aligns better with the class manifold, while ZCA decorrelates the diffusion coordinates to further sharpen prototype matching.

\textbf{Graph Construction and Diffusion Hyperparameters.} Table~\ref{Table3} shows that local scaling consistently improves performance. With $k=10$, accuracy increases from 85.4\% to 86.1\%. On CUB-200-2011 with ResNet-50, using $k=20$ with local scaling, $t=4$, and $L=32$ achieves 87.8\%; the average path (A.P.) length drops from 7.4 to 5.2, and interpretability and calibration metrics also improve. Larger settings such as $k=30$, $t=8$, and $L=64$ yield only 86.1\% with higher cost, so we adopt moderate $k$ and $t$.

\textbf{Nystr\"om Landmarks and Inference Efficiency.} As shown in Table~\ref{Table4}, we evaluate landmark strategies on Stanford Cars with ResNet-50. Per-class K-means with 768 landmarks updated every 20 epochs achieves the best trade-off, reaching 88.9\% accuracy, 2.6\% ECE, and about 5.6 ms latency. Random sampling is faster but less accurate, while global or fixed large-scale landmarks increase latency without clear benefits. Periodic updates and moderate landmark counts offer a good balance between accuracy and efficiency.

\section{Conclusion}
\label{sec:prior}

This paper presents GeoProto, a geodesic prototype learning paradigm for interpretable and fine-grained recognition. It aligns similarity with intrinsic feature geometry on class conditional manifolds and preserves case-based reasoning with strong calibration. Across two benchmark datasets with diverse backbones it improves accuracy while remaining efficient. We believe GeoProto inaugurates a geometry aware paradigm that grounds similarity in manifold geodesics, enabling seamless adoption while preserving case-based reasoning and steadily raising the reliability of predictions.



\bibliographystyle{IEEEbib}
\bibliography{refs}

\end{document}